\documentclass[a4paper, 10pt, conference]{ieeeconf}      
\usepackage{FG2024}

\FGfinalcopy 

\usepackage{amsmath,amssymb,amsfonts}
\usepackage{algorithmic}
\usepackage{graphicx}
\usepackage{textcomp}
\usepackage{xcolor}
\usepackage{algorithmic}
\usepackage{algorithm}
\usepackage{caption}
\usepackage{url}
\usepackage{tabularx,booktabs}

\IEEEoverridecommandlockouts                              
\def\FGPaperID{159} 

\title{\LARGE \bf
Resource-Efficient Gesture Recognition using Low-Resolution Thermal Camera via Spiking Neural Networks and Sparse Segmentation 
}

\author{\parbox{16cm}{\centering
    {\large Ali Safa$^{1,2}$, Wout Mommen$^{1,3}$ and Lars Keuninckx$^1$}\\
    {\normalsize
    $^1$ imec, Leuven, Belgium, $^2$ ESAT, KU Leuven, Belgium, $^3$ ETRO, VUB, Brussels, Belgium}}
    \thanks{This research received funding from the Flemish Government under the “Onderzoeksprogramma Artificiële Intelligentie (AI) Vlaanderen” programme.}
}

\begin{document}

\ifFGfinal
\thispagestyle{empty}
\pagestyle{empty}
\else
\author{Anonymous FG2024 submission\\ Paper ID \FGPaperID \\}
\pagestyle{plain}
\fi
\maketitle

\begin{abstract}

This work proposes a novel approach for hand gesture recognition using an inexpensive, low-resolution ($24 \times 32$) thermal sensor processed by a Spiking Neural Network (SNN) followed by Sparse Segmentation and feature-based gesture classification via Robust Principal Component Analysis (R-PCA). Compared to the use of standard RGB cameras, the proposed system is insensitive to lighting variations while being significantly less expensive compared to high-frequency radars, time-of-flight cameras and high-resolution thermal sensors previously used in literature. Crucially, this paper shows that the innovative use of the recently proposed Monostable Multivibrator (MMV) neural networks as a new class of SNN achieves more than one order of magnitude smaller memory and compute complexity compared to deep learning approaches, while reaching a top gesture recognition accuracy of $93.9\%$ using a 5-class thermal camera dataset acquired in a car cabin, within an automotive context. Our dataset is released for helping future research. 

\end{abstract}


\section*{Supplementary Material}
Our dataset can be downloaded at: 

\url{http://tinyurl.com/4purd2e3} 

\section{Introduction}
Hand gesture recognition is a popular domain of research for human-machine interfacing and has been studied in many different contexts, from touchless interaction with smartphones and computers \cite{9667068, 8756601, 8756576} to in-car driver monitoring applications \cite{8373883, 9320259, 8756566}. 

In addition, different sensor modalities have been investigated in the literature, from RGB and depth cameras \cite{9667068} to high-frequency radars and high-resolution thermal camera sensors \cite{9320161}. Since standard RGB cameras suffer from high sensitivity to lighting conditions \cite{9606515, 10160681}, the use of more \textit{expensive} sensors such as high-frequency radars \cite{9718541, 9666988}, as well as \textit{high-resolution} thermal \cite{8373883} and time-of-flight (ToF) cameras \cite{9667068} has been investigated in scenarios where the gesture detection system must be robust to lighting conditions caused by day-night variations, such as for \textit{in-car driver gesture recognition}, which is the scenario considered in this paper. 

\begin{figure}[!t]
\centering
      \includegraphics[scale = 0.55]{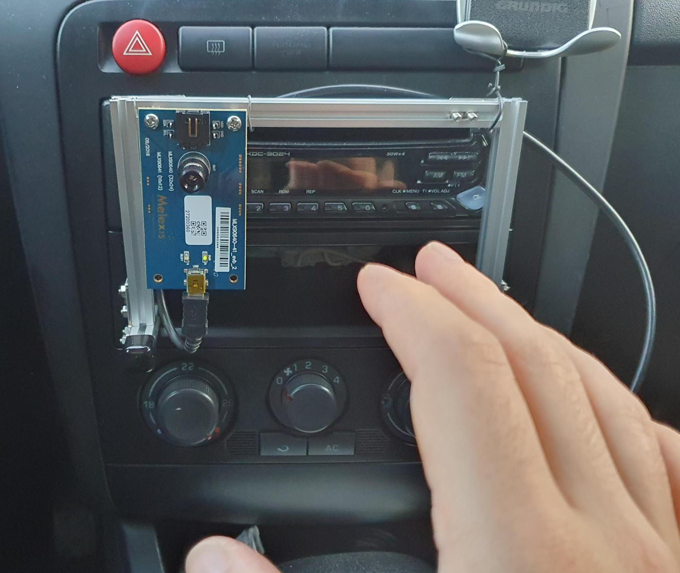}
      \caption{\textit{\textbf{Thermal gesture sensing setup.} A $24 \times 32$ pixel MLX90640 thermal camera is mounted in a car cabin and used to acquire data for the study of gesture recognition.
      }}
      \label{setupthermal}
 \end{figure}

Since radars and ToF cameras can be expensive \cite{9150613}, this work investigates the use of an inexpensive low-resolution thermal sensors for in-cabin driver gesture recognition (see Fig. \ref{setupthermal}), as a sensing modality not affected by day-night lighting variation. 

Crucially, we investigate a complementary approach to previously-proposed deep learning systems such as \cite{9150613} (which uses deep Temporal Convolution Networks), by proposing a \textit{first-of-its-kind} thermal gesture recognition system using an event-based or \textit{spiking} neural networks (SNN) \cite{9320225}. The SNN acts as a highly memory- and compute-efficient \textit{wake-up} system for a downstream sparse segmentation and feature-based classification pipeline using Robust Principal Component Analysis (R-PCA) \cite{10.1145/1970392.1970395}. As SNN model, we use the recently-proposed Monostable Multivibrator (MMV) Neural Networks as a fully-binarized and highly compute- and area-efficient SNN design approach \cite{KEUNINCKX2018224, Keuninckx_2023}.

The contributions of this paper are the following:
\begin{enumerate}
    \item We acquire a novel dataset for the study of in-car gesture recognition via low-resolution thermal camera data (our dataset is released as supplementary material).
    \item Using our dataset, we present a novel memory- and compute-efficient approach for gesture recognition using the recently proposed MMV event-based neural network followed by R-PCA segmentation and feature-based gesture classification. 
    \item We show that our proposed approach significantly outperforms by more than \textit{one order of magnitude} the use of deep learning methods in terms of memory and computational costs, while achieving comparable accuracy.
\end{enumerate}

This paper is organized as follows. Section \ref{background} provides background theory about MMVs and R-PCA segmentation. Section \ref{methods} presents our dataset acquisition and gesture classification methods. Section \ref{experimental} provides our experimental results. Finally, conclusions are provided in Section \ref{conclusions}. 

\section{Background}
\label{background}
\subsection{Network of Monostable Multivibrator Neurons (MMVs)}
\label{mnetsummary}
\begin{figure}[htbp]
\centering
      \includegraphics[scale = 0.47]{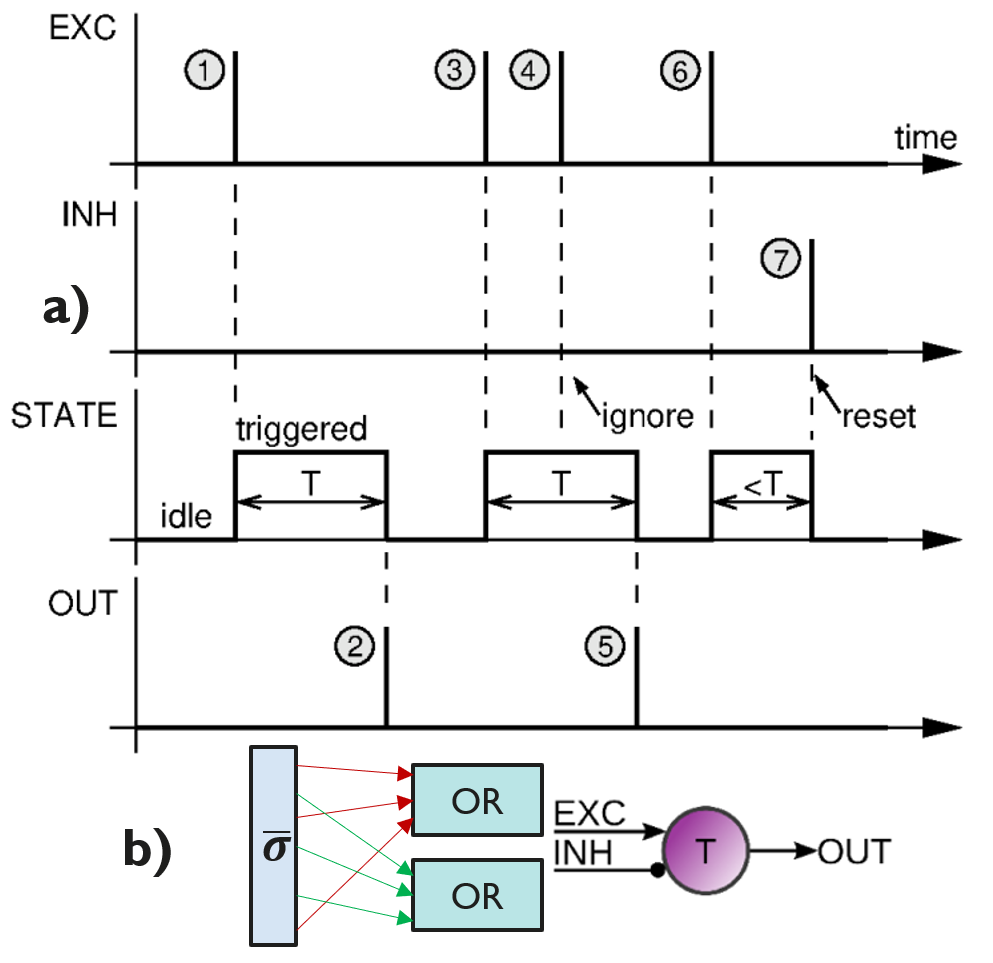}
      \caption{\textit{\textbf{Monostable Multivibrator Neuron.} a) MMV neuron behavior. b) Single MMV neuron connected to an input spiking vector $\Bar{\sigma}$ via the OR-ing binary weights. }}
      \label{mmv_neuron}
 \end{figure}

Recently, a new approach to SNNs has emerged through the use of event-driven binary neurons behaving as Monostable Multivibrator (MMV) \textit{timers} \cite{KEUNINCKX2018224, Keuninckx_2023}. Fig. \ref{mmv_neuron} a) shows the working principle of the MMV neurons. The neuron has two types of inputs: an excitatory (EXC) and an inhibitory (INH) one. As soon as an input spike is received in the EXC connection, a state counter is triggered which counts up to $T$ time-steps. As soon as the counter reached $T$, the MMV counter is reset to zero and an output spike (OUT) is emitted. On the other hand, if a spike is received in the INH connection while the timer is in the triggered state, the timer is reset to zero and the MMV does not output any spikes. 

Using the event-driven MMV neuron of Fig. \ref{mmv_neuron}, networks of fully-connected or recurrent MMV neurons can be built. MMV networks have the additional particularity that all synapses or weight connections are strictly binary: given a $d$-dimensional vector of spikes $\Bar{\sigma}$, each entry $i$ in $\Bar{\sigma}$ is either connected to the EXC input, either to the INH input or not connected at all to the downstream MMV neuron. In addition, the weight-input inner product found in conventional deep neural networks is replaced by a simple logical OR operation between the incoming spikes that are connected to either the EXC or the INH input (see Fig. \ref{mmv_neuron} b). Therefore, during training \cite{Keuninckx_2023}, the MMV network essentially learns what input spike source $\sigma_i$ should be connected to which downstream MMV neuron and with which type of connection (EXC or INH). In addition to learning the binary EXC and INH connections, the integer timer period $T_j$ of each MMV $j$ is learned as well during training \cite{Keuninckx_2023}.

Similar to deep learning, training is carried out using \textit{backprop} with two additional particularities: a) since MMV neurons output spikes and the derivative of a spike in function of the MMV state counter value is ill-defined, this derivative is replaced with a Gaussian function (\ref{surrogate}) acting as a surrogate gradient (see \cite{8891809} for an overview of surrogate gradients in SNNs); b) since the weights of the MMV network are strictly binary, a quantization-aware training procedure is followed which slowly binarizes the weights to either EXC, INH or no connections. We refer the reader to \cite{Keuninckx_2023} for additional details behind MMV network training. 
\begin{equation}
    \text{OUT}^{'}(\text{STATE}) \approx \frac{1}{ \sqrt{2 \pi} } e^{-2 \text{ STATE}^2}
    \label{surrogate}
\end{equation}
Crucially, MMV networks have been shown to be more than one order of magnitude more hardware-efficient compared to traditional SNNs due to their drastic simplification in both neuron model and binary weight connections ($\sim 1$pJ vs. $\sim 200$nJ of energy consumption on MNIST compared to state-of-the-art SNN implementations) \cite{Keuninckx_2023}. Therefore, the compute- and memory-efficient aspect of MMVs justifies their use for our thermal gesture recognition scenario, where our aim is to drastically reduce the hardware footprint compared to conventional deep learning models. 




\subsection{Robust Principal Component Analysis (R-PCA)}

Robust Principal Component Analysis (R-PCA) \cite{10.1145/1970392.1970395} is concerned with the recovery of a low-rank signal matrix $L$ affected by a sparse outlier matrix $S$ with arbitrary amplitude, from the $n_1 \times n_2$ measurement matrix $M$, with:
\begin{equation}
    M = L + S
    \label{sourcesmixed}
\end{equation}

By defining the nuclear norm $||A||_{*} = \sum_i \sigma_i(A)$ as the sum of the singular values of the matrix $A$ and $||A||_1 = \sum_{ij} |A_{ij}|$ as the $l_1$ norm of $A$, it can be shown that the recovery of both $L$ and $S$ from $M$ in (\ref{sourcesmixed}) can be done by solving the following optimization problem \cite{10.1145/1970392.1970395}:
\begin{equation}
    L, S = \arg \min_{L,S} ||L||_{*} + \lambda ||S||_1 \hspace{3pt} \text{s.t} \hspace{3pt} M = L + S
    \label{rpcaopt}
\end{equation}
where $\lambda$ is the sparsity defining hyper-parameter 
\cite{10.1145/1970392.1970395}.

\begin{figure*}[!t]
\centering
    \includegraphics[scale = 0.58]{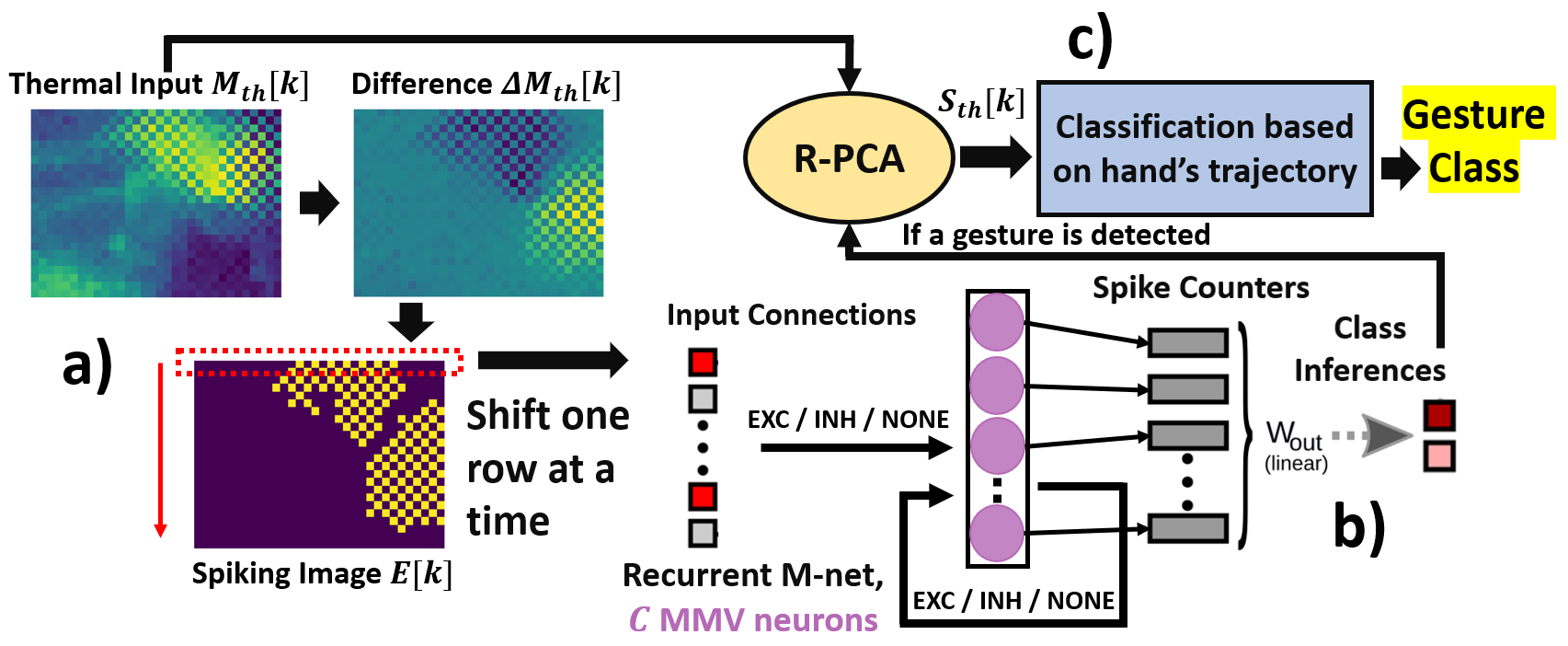}
    \caption{\textit{\textbf{MMV - R-PCA System Overview} a) Pre-processing Eq. \ref{scale}-\ref{spike_convert}  b) Gesture presence detection c) Gesture classification. } }
    \label{systembig}
\end{figure*}

In practice, given a measurement matrix $M$ (such as $N_c$ successive thermal camera data frames), (\ref{rpcaopt}) can be solved using the algorithm for Principal Component Pursuit by Alternating Directions provided in \cite{10.1145/1970392.1970395}, with an open \textit{Python} implementation provided in \cite{rpcacode}. In this paper, we use R-PCA to segment the thermal camera images in order to isolate the \textit{sparse} moving hand in $S$ from the near-static \textit{dense} background in $L$ (see Fig. \ref{systembig} c). After this segmentation, the trajectory of the hand performing the gesture can be tracked and used to classify the gesture type (see Section \ref{classsection}). 

\section{Sensor Processing Methods}
\label{methods}
\subsection{Data Acquisition and Sensing Setup}
\begin{figure}[htbp]
\centering
      \includegraphics[scale = 0.32]{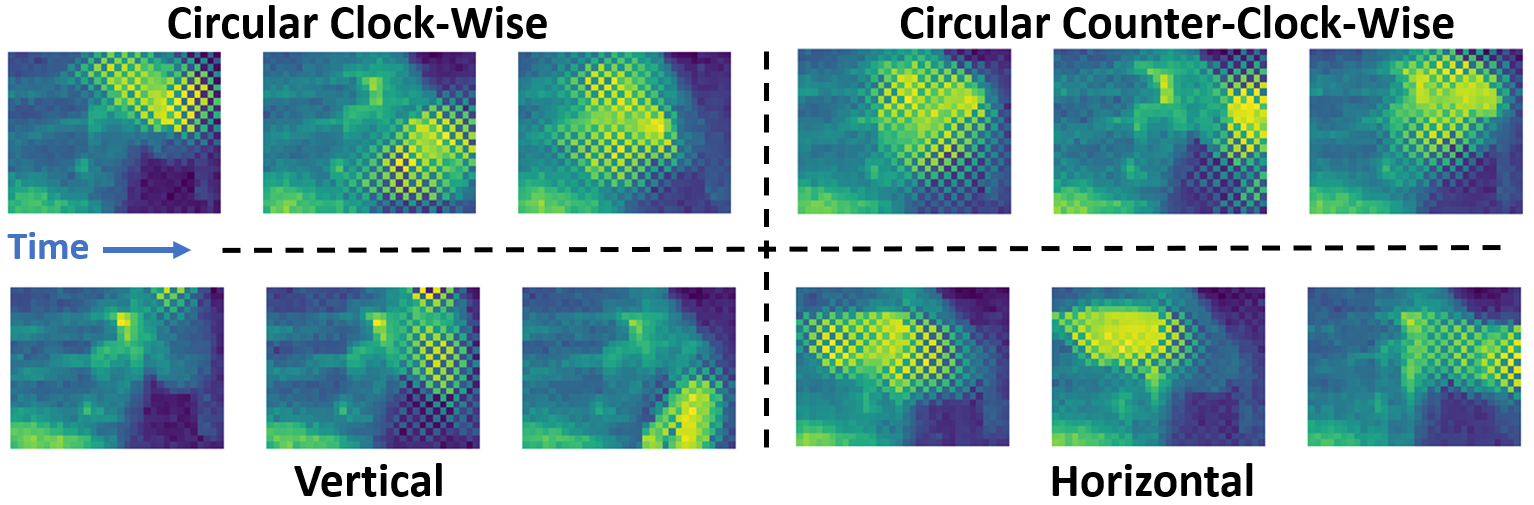}
      \caption{\textit{\textbf{Examples of gestures acquired in our dataset.}  }}
      \label{allgest}
 \end{figure}

We mount a MLX90640 $24 \times 32$ thermal camera inside a car cabin facing the driver seat (see Fig. \ref{setupthermal}) and use it to collect data at 8 frames per second in both day and night times. The in-cabin changes in temperature caused by the day to night shift adds a challenge to this use case. We consider a 5-class recognition problem composed of: i) \textit{circular clock-wise}, ii) \textit{circular counter-clock-wise}, iii) \textit{vertical up-down}, iv) \textit{horizontal right-left} and v) \textit{no gesture} (background and driver's movements). Table \ref{dataacquisitions} lists all data sequences acquired in this work and Fig. \ref{allgest} shows gesture acquisition examples from our dataset.
\begin{table}[htbp]
\centering
\begin{tabularx}{0.47\textwidth}{@{}l*{1}{c}c@{}}
\toprule
Acquisition Name  & Gesture Type   & Frames \\ 
\midrule
no-gesture-\{m,n\}     &  no gestures          &   1853, 1062   \\ 
all-gesture-\{m,n\}     &  all gestures           &  2631, 1109    \\ 
cirCW-gesture-\{m,n\}     &  circular clock-wise          &  824, 727   \\ 
cirCCW-gesture-\{m,n\}    &  circular counter-clock-wise          &   828, 748   \\ 
vert-gesture-\{m,n\}     &  vertical         &     853, 766 \\ 
hor-gesture-\{m,n\}     &  horizontal          &   790, 769   \\ 

\bottomrule
\end{tabularx}
\caption{\textit{\textbf{Gesture data acquired in this work.} In the acquisition name, the suffix -m denotes the morning-noon daytime while the suffix -n denotes night time. The acquisition length is reported accordingly.}}
\label{dataacquisitions}
\end{table}

\subsection{Proposed Gesture Recognition Pipeline}
\label{proposedpipeline}
Our proposed thermal gesture recognition pipeline is a modular system composed of two sub-systems: a) a wake-up system using a very compute-efficient binary MMV network, which is responsible for detecting if any gesture is being performed by the driver; and b) a gesture classification system based on on-line R-PCA thermal camera segmentation and hand-crafted feature extraction. Fig. \ref{systembig} provides a complete overview of the proposed modular system. The MMV-based wake-up sub-system enables a drastic saving in energy consumption by monitoring when an actual gesture is performed (binary detection) and waking up the more compute-expensive R-PCA scene segmentation pipeline only when needed. Then, the R-PCA segmentation filters the environment-dependent background clutter slowly changing through time (i.e., the low-rank contribution $L$ in Eq. \ref{sourcesmixed}) and isolates the pixels corresponding to the fast hand gesture movements as the matrix $S$ in (\ref{sourcesmixed}). Thus, the R-PCA segmentation helps making our system robust to changes in the background, greatly helping model generalization. The trajectory of the segmented hand movement returned in $S$ can then be directly classified using robust hand-crafted features (see Section \ref{classsection}).

\subsubsection{MMV gesture detection}
We use a recurrent MMV network with $C$ neurons followed by a spike counter (which returns the spike count at the output of each MMV neuron after the $T$ MMV execution time steps) and a 2-class logistic regression layer (composed of a $C \times 2$ weight matrix $W_{out}$ followed by $2$ sigmoid neurons) which classifies the spike count result in order to detect if a gesture is being initiated by the user (see the MMV \textit{architecture} in Fig. \ref{systembig} b). As soon as the MMV network detects the execution of a gesture, the R-PCA classification pipeline described in Section \ref{classsection} is woken up.

As input data to the MMV network, a sliding window containing the past $N_c$ thermal frames (flatten as 1D vectors) at time-step $k$ is aggregated as the matrix $M_{th}[k]$ with dimensions $N_c \times (h_{th} w_{tw})$, where the sensor height and width are respectively $h_{th} = 24$ and $w_{tw} = 32$ in Fig. \ref{setupthermal}. Then, the thermal camera frames are normalized to a range of pixel values between $[0, 1]$ using:
\begin{equation}
    M_{th}[k] \xleftarrow{} \frac{M_{th}[k] - \min(M_{th}[k])}{\max(M_{th}[k]) - \min(M_{th}[k])}
    \label{scale}
\end{equation}

After normalization in (\ref{scale}), the thermal frames are differentiated in time in order to capture the \textit{change} in each pixel, as conventionally done with event-driven neural networks \cite{10160681}:
\begin{equation}
    \Delta M_{th}[k] = M_{th}[k] - M_{th}[k-1]
    \label{delta_step}
\end{equation}

Due to differentiation (\ref{delta_step}), the resulting dimension becomes $(N_c - 1) \times (h_{th} w_{tw})$. Then, $\Delta M_{th}[k]$ is transformed into MMV-compatible spiking (binary) data by thresholding the absolute change in pixel intensity against a threshold $\theta_s$:
\begin{equation}
    E_{ij}[k] = 1 \hspace{3pt} \text{if} \hspace{3pt} |\Delta M_{th}[k]| \geq \theta_s  \hspace{3pt} \text{else} \hspace{3pt} E_{ij}[k] = 0 \hspace{3pt} \forall i,j
    \label{spike_convert}
\end{equation}

Finally, the $(N_c - 1) \times (h_{th} w_{tw})$ spiking matrix $E[k]$ is reshaped to $(N_c - 1)h_{th} \times w_{tw}$ and fed row by row into the MMV network, with input dimension $w_{tw}$ during $T = (N_c - 1)h_{th}$ MMV execution time steps. Fig. \ref{systembig} a) shows how the thermal data is fed into the MMV network for gesture detection. The MMV network is trained through backprop following the method described in Section \ref{mnetsummary}, and used to detect the execution of a gesture by the user (see Fig. \ref{systembig} b).

\subsubsection{R-PCA segmentation and feature-based classification}
\label{classsection}
As soon as the MMV network detects the start of a gesture, the R-PCA and classification is woken up. The collection of $N_c$ consecutive thermal camera frames $M_{th}[k]$ at time-step $k$ previously used by the MMVs to detect the presence of a gesture in Fig. \ref{systembig} a) is decomposed by R-PCA into a low-rank contribution $L_{th}[k]$ containing the quasi-static background, and a sparse contribution $S_{th}[k]$ containing non-zero pixels corresponding to the faster gesture movement (see Fig. \ref{systembig} c). The centroid location of the hand performing the gesture is derived from the sparse image $S_{th}[k]$ as the center coordinate $[x_k, y_k]$ of the hand blob \cite{9667069}, and noted $\Bar{p}_c[k] = [x_k, y_k]$. Then, the hand is tracked throughout time $k$ by low-pass filtering the evolution of $\Bar{p}_c[k]$ to alleviate noise and registering the low-pass-filtered coordinate $\Bar{p}_c^l[k]$ as a new point in the track history:
\begin{equation}
    T = \{\Bar{p}_c^l[k], \forall k = k^* - L,...,k^*\}
    \label{track_eq}
\end{equation}
where $k^*$ is the current time step, $L$ is the track length and:
\begin{equation}
    \Bar{p}_c^l[k+1] = \beta \Bar{p}_c^l[k] + (1 - \beta) \Bar{p}_c[k+1]
    \label{lowpass}
\end{equation}
where $\beta \in [0,1]$ is the decay parameter setting the strength of the low-pass filter.

Finally, the gesture class is derived from the track history $T$ in (\ref{track_eq}) as follows:
\begin{enumerate}
    \item If the \textit{difference} between maximum extend in the $x$ and $y$ directions $D_x = \max(T_x) - \min(T_x)$ and $D_y = \max(T_y) - \min(T_y)$ is smaller than a threshold $\theta_{c,1}$: $|D_x - D_y| < \theta_{c,1}$; and the minimum diameter is larger than a threshold $\theta_{c,2}$: $\min(D_x,D_y) > \theta_{c,2}$, the gesture is classified as \textbf{\textit{circular}}.

    The circular gesture is considered \textbf{\textit{clock-wise}} if the angular evolution of the track $T_{\phi} = \arctan \frac{T_y}{T_x}$ is decreasing and considered \textbf{\textit{counter-clock-wise}} in the opposite case.

    \item Else, we test if the gesture is \textit{vertical} or \textit{horizontal} by first measuring the variances of the track evolution along the x-axis and along the y-axis: $\sigma_x^2$ and $\sigma_y^2$. If the variance along the y-axis is larger than along the x-axis $\sigma_y^2 > \sigma_x^2$, the gesture is registered as \textbf{\textit{vertical}}.

    \item Finally, if the inverse is true: $\sigma_x^2 >  \sigma_y^2$, the gesture is registered as \textbf{\textit{horizontal}}.

\end{enumerate}

The first case effectively verifies circularity of the track by checking if diameters along the x- and y- are similar (as it is the case for a circle), and also checks whether the circle is larger than a certain minimum diameter $\theta_{c,2}$. If the first case is not verified, the vertical or horizontal nature of the gesture is identified by checking along which axis changes occur the most by, measuring the variances of $T_x$ and $T_y$.

\section{Experimental Results}
\label{experimental}
We evaluate the accuracy of the proposed system as follows. We first train the MMV wake-up system using only the \textit{no-gesture-m} and \textit{all-gesture-m} acquisitions in Table \ref{dataacquisitions}, randomly partitioned into a training and validation set following a 70\%-30\% split. The training is done using the Adam optimizer with learning rate $\eta = 5\times 10^{-3}$ and batch size $32$ during 50 epochs. MMV network binarization starts from epoch 10 and ends at epoch 25. After binarization, the model keeps on being trained for an addition 25 epochs and MMV accuracy is assessed on the validation set after each epoch. The MMV network parameters with maximum validation accuracy is kept as the final model. Training is carried for three MMV network configurations using respectively 125, 250 and 500 neurons. In addition, Table \ref{hyperparams} lists the various MMV and R-PCA hyper-parameters used in our experiments (tuned manually for reducing compute complexity while not sacrificing accuracy).
\begin{table}[htbp]
\begin{center}
\begin{tabular}{|c|c|c|c|c|c|c|c|c|}
\hline
$N_c$ & $h_{tw}$ & $w_{tw}$ & $L$ & $\beta$  &  $\theta_{s}$ &  $\theta_{c,1}$ & $\theta_{c,2}$ & $\lambda$ \\
\hline
$5$ & $24$ & $32$ & $10$ & $0.5$ & $0.2$ & $5$ & $5$ & $0.05$ \\
\hline
\end{tabular}
\caption{\textit{\textbf{System hyper-parameters.}}}
\label{hyperparams}
\end{center}
\end{table}

Then, the MMV network is integrated into our proposed modular gesture recognition system described in Section \ref{proposedpipeline} (see Fig. \ref{systembig}) and accuracy is assessed using all other acquisitions of Table \ref{dataacquisitions}. This assessment approach makes the problem challenging since 
the night-time data is not used during the model training, but rather only used during testing. These aspects enable a challenging and fair assessment of model generalization to sensing contexts outside its training data (i.e., daytime vs. night-time).   

The performance of the proposed MMV - R-PCA system is reported in Table \ref{performancetabel} and compared against the state-of-the-art 9-class deep learning method proposed in \cite{9150613} using a Temporal Convolution Network (TCN) with the same low-resolution sensor used in this work (see Fig. \ref{setupthermal}). We also compare our proposed system against a baseline MMV-only classifier (using 500 MMV neurons), \textit{without} the R-PCA classification pipeline of Section \ref{classsection}.
\begin{table}[htbp]
\centering
\begin{tabularx}{0.47\textwidth}{@{}l*{2}{c}c@{}}
\toprule
Model   & Accuracy [\%] & Parameters (kB) & Avg. FLOPS \\  
\midrule
TCN f128 \cite{9150613} &  \textbf{95.9} &    12550 &  622.4M\\
TCN f64 \cite{9150613} &  93 &    1090 &  34.01M \\
\midrule
Ours (125)      &  92.6 &    \textbf{36.7} &  \textbf{129k + 1.7M}\\ 
Ours (250)    &  93.56   &    50.5 &  257k + 1.7M     \\ 
Ours (500)  &  93.9 &    101.4   &    513k + 1.7M   \\ 
\midrule
\textit{MMV only (500)}     &  \textit{85.4} &    \textit{76.7} &  \textit{1.28M}\\

\bottomrule
\end{tabularx}
\caption{\textit{\textbf{Accuracy, Parameter Utilization and FLOPS}. The average FLOPS is computed assuming a maximum user gesture execution rate of 1 gesture per minute. For our proposed systems (with 125, 250 or 500 neurons), the Avg. FLOPS is reported as "MMV + R-PCA". We also train an MMV-only setup (without R-PCA) as baseline model.}}
\label{performancetabel}
\end{table}

Table \ref{performancetabel} shows that our proposed MMV - R-PCA system \textit{significantly outperforms} the use of TCN in terms of parameter utilization (memory) and average Float Operation Per Seconds (FLOPS)\footnote[1]{The FLOPS for the Singular Value Decomposition (SVD) executed for R-PCA \cite{rpcacode} is computed by considering the formula provided by \cite{trefethen97} for the number of operations in SVD: $n_{svd} = 2N_c(h_{th}w_{th})^2 + 11(h_{th}w_{th})^3$ and considering a maximum number of $100$ iterations in R-PCA \cite{rpcacode}.}. In addition, Table \ref{performancetabel} demonstrates that our proposed R-PCA-based classification approach of Section \ref{classsection} leads to a significant gain of $+8.5\%$ in accuracy compared to the MMV-only baseline classifier. On the other hand, the TCN in \cite{9150613} achieves superior accuracy to our system, with the ability to recognize more classes than ours (9 vs. 5 classes), while suffering from more than one order of magnitude more memory and compute complexity. 

\section{Conclusion}
\label{conclusions}

This paper has presented a memory- and compute-efficient gesture recognition system using low-resolution thermal camera data. After acquiring a novel thermal camera dataset for gesture recognition, a first-of-its-kind system using the recently-proposed MMV approach followed by conventional R-PCA and feature-based gesture classification has been presented. Thanks to its use of an ultra-low-complexity MMV wake-up system, it was shown that the proposed gesture detector achieves close accuracy to state-of-the-art deep learning systems while consuming more than one order of magnitude less memory and compute power. Hence, the proposed approach is well-suited for application cases where accuracy can be slightly traded off for a significant gain in memory and compute efficiency. Finally, we released our dataset for helping future research.
\vspace{12pt}


{\small
\bibliographystyle{ieee}
\bibliography{egbib}
}

\end{document}